\documentclass[11pt]{article}
\usepackage{emnlp2021}
\usepackage{times}
\usepackage{latexsym}
\usepackage{tabularx}

\usepackage{multirow}
\usepackage{comment}
\usepackage{sidecap}
\usepackage{graphicx}
\usepackage[export]{adjustbox}
\usepackage[colorinlistoftodos]{todonotes}

\usepackage{latexsym}
\usepackage{soul}
\usepackage{amsfonts}
\usepackage{amsmath}
\usepackage{mathtools}
\usepackage{amssymb}
\usepackage{mathptmx}
\usepackage{nicefrac}
\usepackage{booktabs} % Pretty tables
\usepackage{makecell} % cell line breaks..
\usepackage{graphicx}
\usepackage{float}
\usepackage[inline]{enumitem}
\usepackage[font=small]{subcaption}
\usepackage[font=small]{caption}
\usepackage{xargs} % commandx
\usepackage[autostyle]{csquotes}  

\usepackage{hyperref}
\hypersetup{
    colorlinks=true,
    linkcolor=blue,
    filecolor=magenta,      
    urlcolor=blue,
}

\usepackage[T1]{fontenc}

\usepackage{color}
\usepackage{soul}
\usepackage{todonotes}

\definecolor{pigment}{rgb}{0.2, 0.2, 0.6}

\definecolor{blue}{RGB}{0, 93, 170}			
\definecolor{darkgreen}{HTML}{3bb35b}

\definecolor{darkyellow}{HTML}{f0b71d}

\newcommand{\ignore}[1]{}
 \usepackage{multirow}
 \usepackage{float}
 \usepackage{physics}

 \usepackage{bbm}
 
 \usepackage{amsmath}
\DeclareMathOperator*{\argmax}{arg\,max}

\title{Calibration of Machine Reading Systems at Scale}

\newcommand{\Read}{\textsc{Reader }}
\newcommand{\Ret}{\textsc{Retriever }}

\author{Shehzaad Dhuliawala$^{1}$,  Leonard Adolphs$^{1}$, Rajarshi Das$^{2}$, Mrinmaya Sachan$^{1}$ \\
   \\
  $^{1}$ETH Z\"urich,
  $^{2}$University of Massachusetts Amherst \\
  \texttt{\{firstname.lastname\}@inf.ethz.ch}}

\begin{document}

\maketitle
\begin{abstract}
    \label{sec:abstract}
    
In typical machine learning systems, an estimate of the probability of the prediction is used to assess the system's confidence in the prediction.
This confidence measure is usually uncalibrated; i.e.\ the system's confidence in the prediction does not match the true probability of the predicted output.
In this paper, we present an investigation into calibrating open setting machine reading systems
such as open-domain question answering and claim verification systems.
We show that calibrating such complex systems which contain a discrete retrieval and deep reading components is challenging and current calibration techniques fail to scale to these settings. 
We propose simple extensions to existing calibration approaches that allow us to adapt these calibrators to these settings.

Our experimental results reveal that the joint calibration of the retriever and the reader outperforms the reader calibrator by a significant margin. We also show that the calibrator can be useful for selective prediction, e.g., when question answering systems are posed with unanswerable or out-of-the-training distribution questions.
\end{abstract}

\section{Introduction}
    \label{sec:introduction}
    
With recent advances in machine reading, there has been a surge of interest in practical applications of the technology such as open-domain question answering \cite{karpukhin2020dense, lee-etal-2019-latent} and claim verification \cite{thorne2018fact}. Due to various scale limitations in practical settings, these systems are seldom trained end-to-end. Such systems typically make use of a \Ret alongside a \Read -- the evidence is first retrieved from a large corpus and is then used by a machine reading model to provide an answer.

As these systems are increasingly being deployed in the real world, it is important that they are not only accurate but also trustworthy.
A way to make these systems trustworthy is to indicate when they are likely to be incorrect by providing a calibrated \texttt{confidence} measure in addition to the prediction.

A naive solution for this is to use the system's output probability as the confidence. 
However, this confidence score is often \texttt{uncalibrated}  \cite{kuleshov2015calibrated,guo2017calibration}; i.e.\ it is not representative of the true correctness likelihood.\footnote{For a perfectly calibrated system, given 100 answer predictions, each with a confidence of 0.7, we expect that 70 should be correct.}

Previous work \cite{jiang2020can,jagannatha2020calibrating,desai2020calibration} has shown that large language models especially suffer from miscalibration. Thus, several methods have been proposed to calibrate language models based on gradient-based calibration methods such as temperature scaling \cite{guo2017calibration} and feature-based forecasters \cite{kuleshov2015calibrated}. While gradient-based calibration is intuitive and easy to implement, feature-based forecasters require manual feature engineering.

In this work, we contribute a simple method to calibrate practical \Ret- \Read machine reading pipelines.
These systems typically include a hard retrieval step which makes gradient-based calibration infeasible. 
Thus, we make use of the Gumbel machinery \cite{jang2017categorical,maddison2017concrete}; specifically the Gumbel top-$K$ procedure of \newcite{vieira2014gumbel, xie2019reparameterizable} to obtain a differentiable sampling routine for the retrieval step. This sampler can then be combined with any gradient-based calibration technique such as Platt's scaling.

We conduct experiments on three different models -- a generative and extractive open-domain question answering model and a claim verification model.
We find that calibrating the \Ret and the \Read jointly is better than calibrating only the \Read or the \Ret.
We also show that our approach can produce calibrated scores that can be used to selectively abstain from answering questions that are contrived or ill-posed or questions that are out-of-the-training distribution. 
Finally, we also demonstrate how the calibration of such a system works -- the calibration techniques lower the confidence of the predicted answer when the question is unanswerable or when the retriever is not able to retrieve any relevant evidence for answering the question. 
We perform our analysis on different types of unanswerable questions and show that incorporating the confidence of the \Ret along with the \Read can improve the confidence estimate of the answer.

\section{Preliminaries}    
    \label{sec:open}
    \subsection{Machine Reading at Scale}
\begin{figure}
    \centering
    \includegraphics[width=\linewidth]{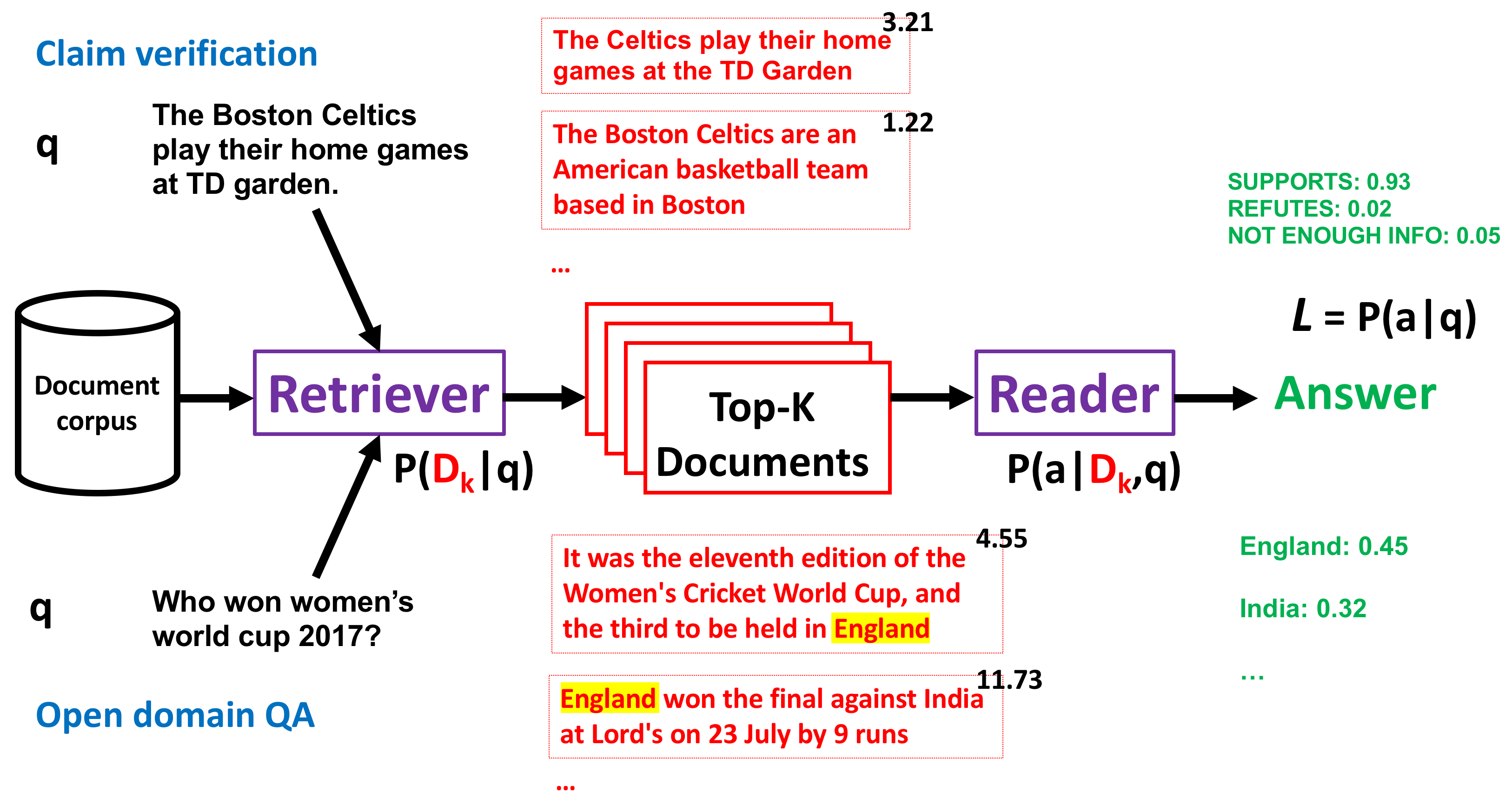}
    \caption{General architecture of the two machine reading systems considered in this paper. a) Claim verification (top half) and b) Open-domain QA (bottom half). The systems follow the same architecture and are composed of a retriever and a reader. Given the query, the retriever retrieves a set of K documents from the corpus along with scores for each of them. The reader then takes these as input and produces the output: a veracity label for claim verification and an answer span for the QA model. This can be seen as a probabilistic model with latent retrieval (\textcolor{red}{$D_k$} shown in red). The goal of this paper is to calibrate the final output probabilities $P(a|q)$.}
    \label{fig:arch}
\end{figure}

Practical real-world machine reading systems such as open-domain question answering systems \cite{chen2017reading} \cite{karpukhin2020dense} \cite{izacard2020leveraging} or claim verification systems \cite{hanselowski2018ukp} rely on an information retrieval (IR) component called a \Ret to reduce the search space over a large corpus of documents.
This smaller set of documents is then passed to a \Read model that reasons over the text and produces an answer. 
This setting, where the \Read is not given labeled documents is referred to, in the literature, as an open-domain setting.

We now proceed to formally define the pipeline for a machine reading system in the open-domain setting. 

Let $\mathcal{D} = \{d_1, \dots, d_N\}$ denote the given corpus of documents. 
Let $q$ denote the user query (a question or a claim).
We denote the answer to the question or the veracity label of the claim as $a$.
The retreiver model takes in $q$ and scores all the documents $d \in \mathcal{D}$ to produce a set of scores:
\begin{align}
    \textsc{Retriever}( d_1, \dots, d_N | q) \longrightarrow S_{d_1}, \dots, S_{d_N} 
\end{align}
This formulation of the \Ret is generic. This allows our method to work with any IR model such as the traditional BM25 model \cite{wiki:bm25} to more modern methods such as Dense Passage Retrieval (DPR) by \citet{karpukhin2020dense}. 

The documents are then sorted based on the scores and the $k$ top-scoring documents are chosen. 
We call this set of top-$K$ documents $D_k$.
$D_k$ is then given to a \Read model which extracts the answer or predicts a veracity label for the claim, $a$.
The \Read can vary depending on the task. For extractive QA, the \Read produces a score for each span ($s_i$) in the documents provided to it.
\begin{align}
    \Read(q, D_k) \longrightarrow S^{Read}(s_i), s_i \in D_k
\end{align}

In claim verification, the \Read produces a score for each veracity label: \textsf{SUPPORTED}, \textsf{REFUTED} or \textsf{NOT ENOUGH INFO}, which indicate whether the claim can be verified by the given set of documents.

\begin{align*}
    \Read(D_k, q) \longrightarrow & S_{\textsf{SUPPORTED}}, \\ & S_{\textsf{REFUTED}}, 
     \\&S_{\textsf{NOT ENOUGH INFO}},
\end{align*}

\subsection{Calibration}

We summarize below the calibration framework \cite{kuleshov2015calibrated} in the context of machine reading. 
Given a query $q$, true output $a$, model
output $\hat a$, and probability $P(\hat a|q)$ calculated over this output, a perfectly calibrated model satisfies the following condition:

\begin{eqnarray}
\mathbb{P}(\hat a = a| P(\hat a|q) = p) = p \hspace{0.2cm}\forall p \in [0,1]
\end{eqnarray}

In simple words, for the confidence estimate $P(\hat a|q)$ to be calibrated, we require that $P(\hat a|q)$ follows the unknown true probability distribution $\mathbb{P}$.

In a multi/binary class setting, a calibrator can be learned to map the output distribution to a calibrated confidence score. However, in a machine reading setting, the space of possible documents retrieved and answers contained in them is usually very large. Thus, we only focus on a specific event set $I(q)$ of interest.
The event set $I(q)$ can be defined using the outputs relevant to the deployment requirements of the machine reading model. In our work, we consider 
all answer candidates in the retrieved set of documents $D_k$: $I(q) = \{a | a \in \argmax\limits_{D_k} P(\hat a|D_k, q)\}$

\subsection{Measuring Calibrated-\textit{ness}}

Calibration can be measured by computing the difference in expectation between confidence scores and accuracies.
%Let P\
\begin{align}
    \mathbb{E}_{P(\hat a|q)} \Big[ \mathbb{P}(\hat a = a | P(\hat a|q) = p) - p \Big]
\end{align}
This is known as expected calibration error (ECE) \cite{naeini2015obtaining}. Practically, ECE is estimated by partitioning the predictions in \textit{M} equally spaced bins $(B_1 \dots B_M)$ and taking the weighted average of the difference between the average accuracy and average confidence of the bins.
\begin{align}
    \textsc{ECE} = \sum_{m=1}^{M} \frac{|B_m|}{n} \left| \textit{acc}(B_m) - \textit{conf}(B_m)\right| 
\end{align}
\subsubsection{Reliability Diagrams}
Another common tool to visualize model calibration is a reliability diagram. A reliability diagram plots sample accuracy as a function of confidence for each bin. If a model is perfectly calibrated, the confidence and accuracy bars should be identical. 

\subsection{Calibration methods}
The general algorithm used for calibrating classification models involves transforming the logits produced by the model. The parameters for this transformation are trained on a held-out calibration set $C = \{(q_i,a_i)\}_{i=1}^N$. This method has been shown to improve the model's ECE without a significant loss in accuracy.
In our work, we use negative log-likelihood (NLL) to tune a model $P(a |q)$ to be a good probability estimate of the output answers:
\begin{eqnarray}
l_\theta = - \sum_{i=1}^N log(P(a_i|q_i))
\end{eqnarray}

ML theory guarantees that NLL is minimized if and only if $P(a_i|q_i)$ recovers the ground-truth conditional distribution $\mathbb{P}(a |q)$.  In the following part of this section, we describe some of these key methods.

\subsubsection{Temperature Scaling}
Temperature scaling \cite{guo2017calibration} is one of the simplest methods for calibration and has been shown to be very effective. 
Temperature scaling allows the logits of the system's output ($Z$) to be scaled by a single temperature value $\tau$. 
This scaling is done before the computation of the softmax.

\vspace{-0.15in}
\begin{align}
    Y = \texttt{softmax}(Z/\tau)
\end{align}
We optimize $\tau$ by maximizing $\mathcal{L}_\theta$ on the dev set.

\subsubsection{Temperature prediction}
The temperature prediction approach \cite{kumar2019calibration} extends temperature scaling to a gradient-based approach. The output logits of the classifier are featurized and passed through an MLP which predicts a temperature value. This temperature value is used to scale the logits.
In contrast to temperature scaling which learns one temperature parameter for each example, in this approach, a new temperature value can be learned for each example.

\begin{align*}
\frac{1}{\tau_i} = \sigma(\texttt{MLP}(Z_i))\\
Y_i = \texttt{softmax}(Z_i/\tau_i)
\end{align*}

\subsubsection{Forecasters}
Forecasters were introduced to calibrate structured prediction models \cite{kuleshov2015calibrated,jagannatha2020calibrating}.
%When the output space is large, the user might be interested in the top output being calibrated. 
The forecaster approach introduces a feature-rich calibration model %called a forecaster. 
that uses various features of the model such as its logits and various uncertainties estimated to predict the confidence score. 
This approach generally only produces a calibrated score over a smaller set of candidate predictions referred to as the interest set $I(.)$
Previous work has successfully used gradient boosted decision trees \textbf{(XGB)} as forecasters.

\section{Calibration of Machine Reading Systems}
    \label{sec:calib}

Previous work has looked at calibration in the aspect of machine reading  \cite{jagannatha2020calibrating,jiang2020can}. 
However, these works do not consider the open setting in which the evidence document for each query is not provided.
We are interested in determining the calibrated probability distribution of the system, $\mathbb{P}(a| q)$.
In the first set of methods, we do this by calibrating the confidence of the model $P (\hat a = a| q)$.
For a machine reading system, 
\begin{align}
     P (\hat a = a| q) = \sum_{D_{k} \in \mathcal{D}} \underbrace{P(D_k| q)}_\text{conf of \Ret} \hspace{-5pt} \times \underbrace{P(\hat a = a | q, D_k)}_\text{conf of \Read}
     \label{eq:MR}
\end{align}

We discuss three possible ways to calibrate $P (\hat a = a| q)$
\paragraph{\textsc{Only reader}}
One way to calibrate $P (\hat a = a| q) $ is to assume that the \Ret is perfectly accurate and perfectly calibrated.
%, i.e., $P(\hat a = a| q) = P(\hat a = a| q)$. 
We refer to his approach in our results as \textsc{Only reader}.
In this approach, we only calibrate $P(\hat a = a | q, D_k)$.
We can use all the previously mentioned calibration approaches for this task. Indeed, this is the approach taken by \cite{jagannatha2020calibrating,jiang2020can}.
For extractive QA, the output logits lie over all the possible text spans, while for fact verification we have a single logit per class. 
In our experiments, we show that this leads to subpar calibration.

\paragraph{\textsc{Individually calibrated}}
We explore another possible approach where we calibrate $P(D_k| q)$ and $P(\hat a = a | q, D_k)$ indivdually using the objectives of the \Ret and \Read individually. 
We refer to this approach as \textsc{Individually calibrated}.
 
This happens in two steps, first the confidence of the retriever, $P(D_k| q)$, is calibrated. The confidence of the retriever is then fixed and the we calibrate the only the confidence of the reader, $P(\hat a = a | q, D_k)$. Finally, the confidence of the systems is computed as $P (\hat a = a| q)$ in Equation \ref{eq:MR}.

We posit that this method results in subpar calibration owing to the \Ret not having gold labels and is calibrated using the less accurate distance supervision objective.

\paragraph{\textsc{Jointly Calibrated}}
Finally, we discuss our approach to calibrate the entire system using the final objective of the system. 
We refer to this approach as \textsc{Jointly Calibrated}.
In this approach, we treat the documents retrieved by the retriever as a latent variable $\textcolor{red}{D_k}$. 

We define our calibration likelihood in eqn. 6 as:
\begin{align}
\mathcal{L}_\theta = \sum_q\sum_{\textcolor{red}{D_{k}} \in \mathcal{D}} P(\textcolor{red}{D_{k}}| q)P(\hat a = a | q, \textcolor{red}{D_{k}})
\end{align}
Clearly, it is infeasible to marginalize over all possible $D_{k}$ (subsets of the corpus of size $k$). Thus, we propose a diffentiable sampler for $D_{k}$:
\begin{align}
\mathcal{L}_\theta = \sum_{D_{k} \sim P_\theta(D_k | q)} P(\hat a = a | q, D_{k})
\end{align}

To make our calibrator differentiable, we apply the Gumbel–softmax trick \cite{maddison2017concrete} and, in particular, its extension to top-K subset selection \cite{vieira2014gumbel,xie2019reparameterizable}. 
The Gumbel-top-$K$ trick generalizes Gumbel–softmax and essentially repeats the Gumbel trick $K$ times until we have a set of the desired size.
We describe the approach further in Appendix \ref{appendix:gumbel-topk}.

\section{Experimental Details}
    \label{sec:exp}

\subsubsection{Open Domain Question Answering}
\paragraph{Extractive}
We test the described calibration techniques on the open domain QA, using the pretrained models from \cite{karpukhin2020dense}. 
We perform our experiments on the Natural Questions (NQ) dataset \cite{kwiatkowski2019natural}.
We randomly split our validation set into two equal parts which we will call \texttt{calib} and \texttt{valid}. We use these splits for training and tuning our calibration models respectively. 
We use the test set of NQ as our test set (\texttt{test}). 
During inference, we use the \Ret to retrieve top 10 documents which are passed to the \Read to extract the answer. 

\paragraph{Generative}
We use the FiD model proposed by \cite{izacard2020leveraging} for our calibration experiments. 
As generative models don't produce a confidence over multiple answers, we use the approach described by \cite{jiang2020can} to generate an interest set.
First we calculate the probabilities of the first generated tokens. We mask out any tokens not in the retrieved passages. Next we, select the top $R$ tokens we find their location in the passages and calculate the probability of all continuing spans up to a certain length (of 10 tokens). We then keep the top-10 scoring spans in our candidate set. 

\subsubsection{Claim Verification}
For the claim verification task, we experiment on the FEVER dataset \cite{thorne2018fever}. 
We use a recently published state-of-the-art model, \cite{liu2020kernel}, in our calibration experiments. 
For every test example, we retrieve 5 sentences that are provided to the claim verification model to ascertain the veracity of the claim.

\subsubsection{Temperature based methods}
For the \Read-\Ret setup we require two temperature parameters \texttt{t1} and \texttt{t2} for the \Ret and \Read respectively. 
We use gradient descent to optimize \texttt{t1} and \texttt{t2} by maximizing $\mathcal{L}_\theta$ on the \texttt{valid} set.
For temperature prediction we add a 2-layer MLP that predicts \texttt{t1} and \texttt{t2} for each example. Once again, the optimization is performed on \texttt{valid}.

\subsubsection{Forecaster}
For our forecaster, we use gradient boosted decision trees. 
We train the model to perform binary classification with the model's accuracy as the objective, i.e., if the model's prediction was correct, we assign a positive label to the example. 
We do not experiment extensively with various features as previous work has done and instead just use the raw logit scores.
Similar to \citet{jagannatha2020calibrating}, we create the interest set of the forecaster by choosing the top-3 predictions of the model, i.e., we choose the top-$3K$ choices of the \Ret over which we evaluate our \Read and choose the top-3 choices.

\subsubsection{Gumbel top-$K$}
For the Gumbel top-$K$ approach required to train the vector scaling and temperature prediction models, we start out with a high temperature value $T_0$ which we linearly decrease to $T_{\infty}$. We treat these parameters as hyperparameters.

\section{Results}
    \label{sec:results}
    
\begin{figure}
    \centering
    \includegraphics[width=\columnwidth]{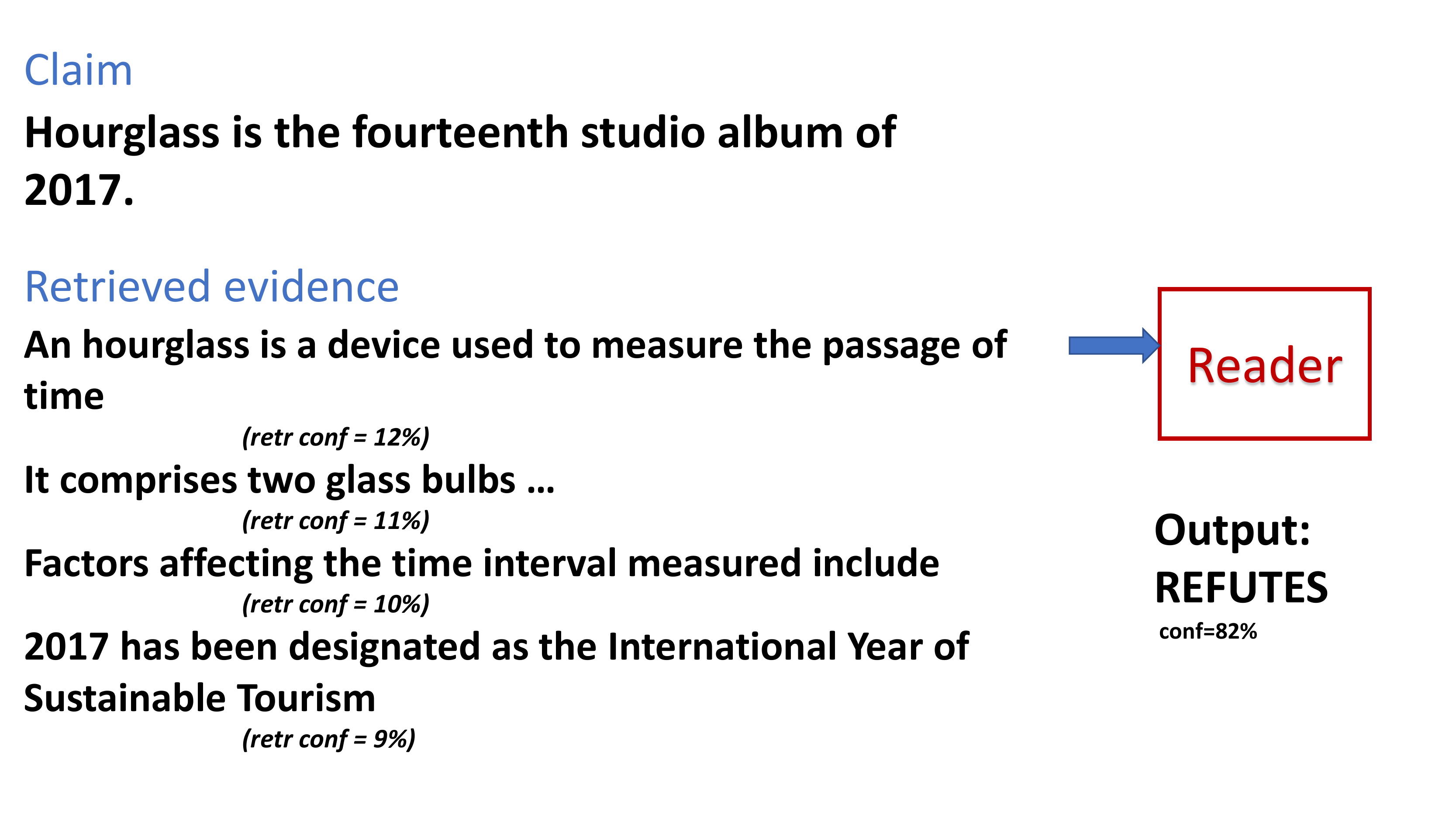}
    \caption{The \Read is highly confident about its prediction, but when we incorporate the confidence of the evidence from the \Ret which can identify that the sentences are irrelevant to the claim, the confidence of the prediction can be better calibrated.}

    \label{fig:bad_reader}
\end{figure}

\begin{figure*}
  \label{fig:reliable}
   \hspace{-2pt}
  \begin{subfigure}[t]{0.3\textwidth}
    \centering	
\includegraphics[trim={2cm 0 1cm 0},clip, keepaspectratio, width=6cm]{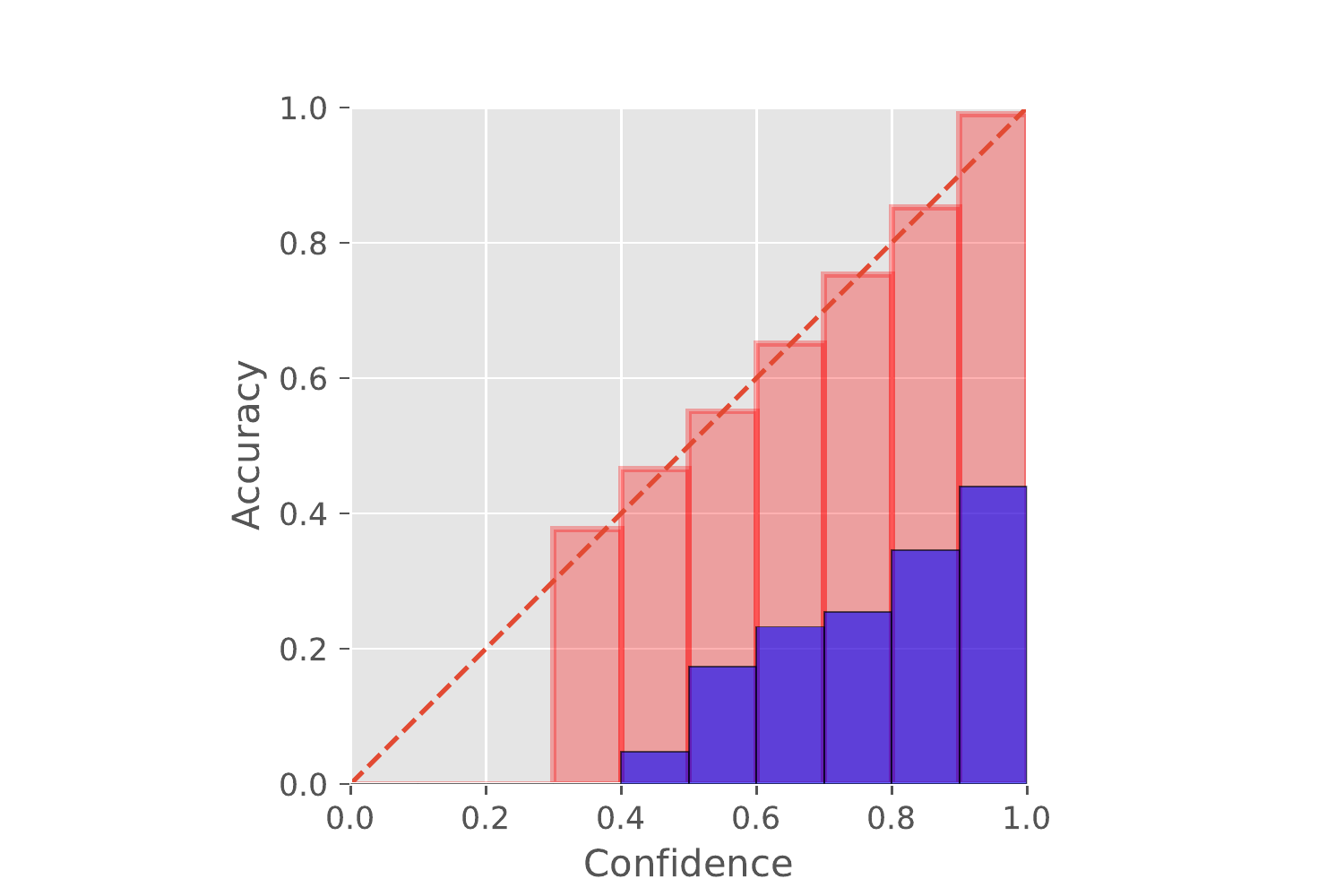}
    
  \end{subfigure}
 \hspace{1pt}
  \begin{subfigure}[t]{0.3\textwidth}
    \centering
    \includegraphics[ trim={2cm 0 1cm 0}, clip, keepaspectratio, width=6cm]{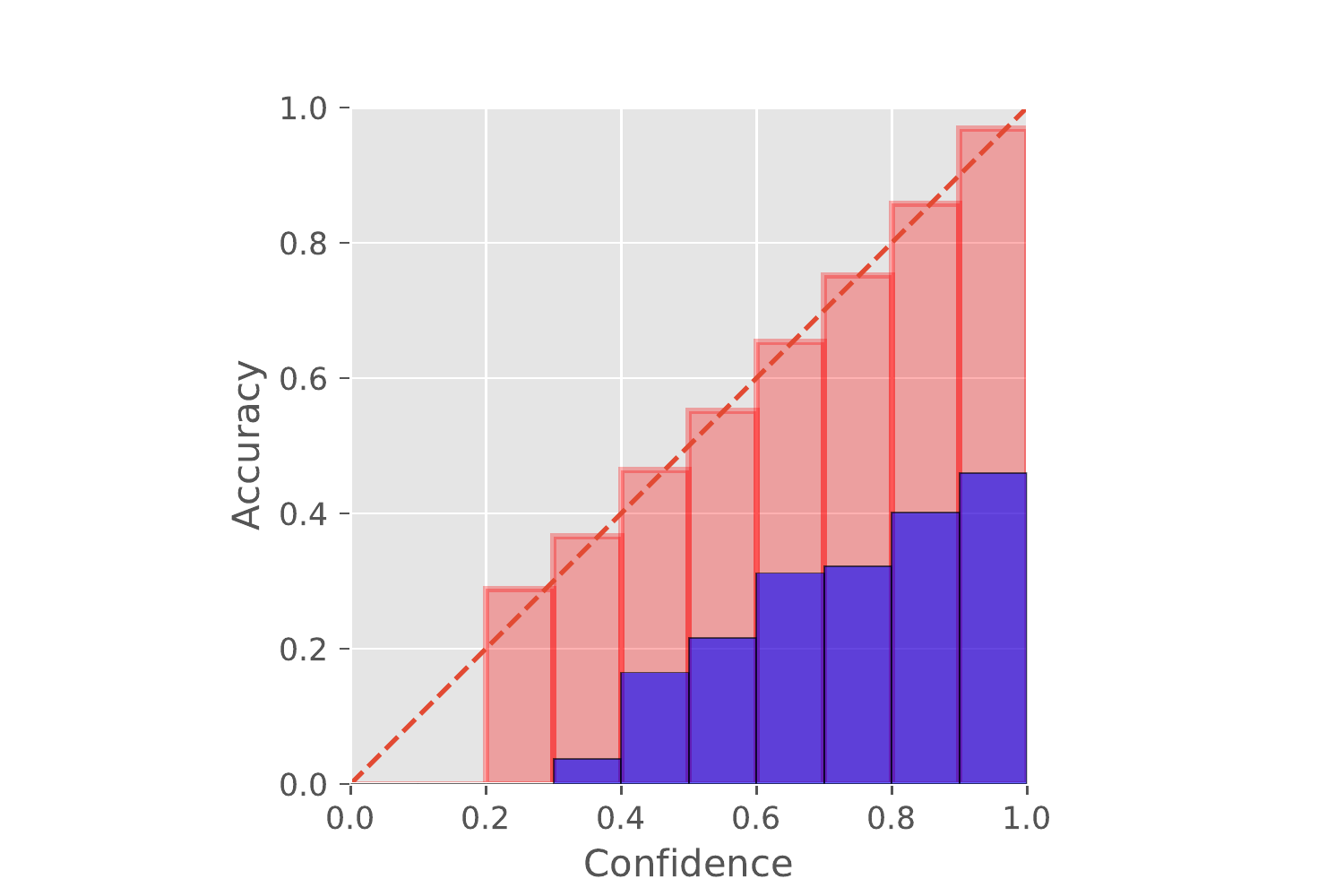}
  \end{subfigure}
 \hspace{1pt}
  \begin{subfigure}[t]{0.3\textwidth}
    \centering
    \includegraphics[trim={2cm 0 1cm 0}, clip, keepaspectratio, width=6cm]{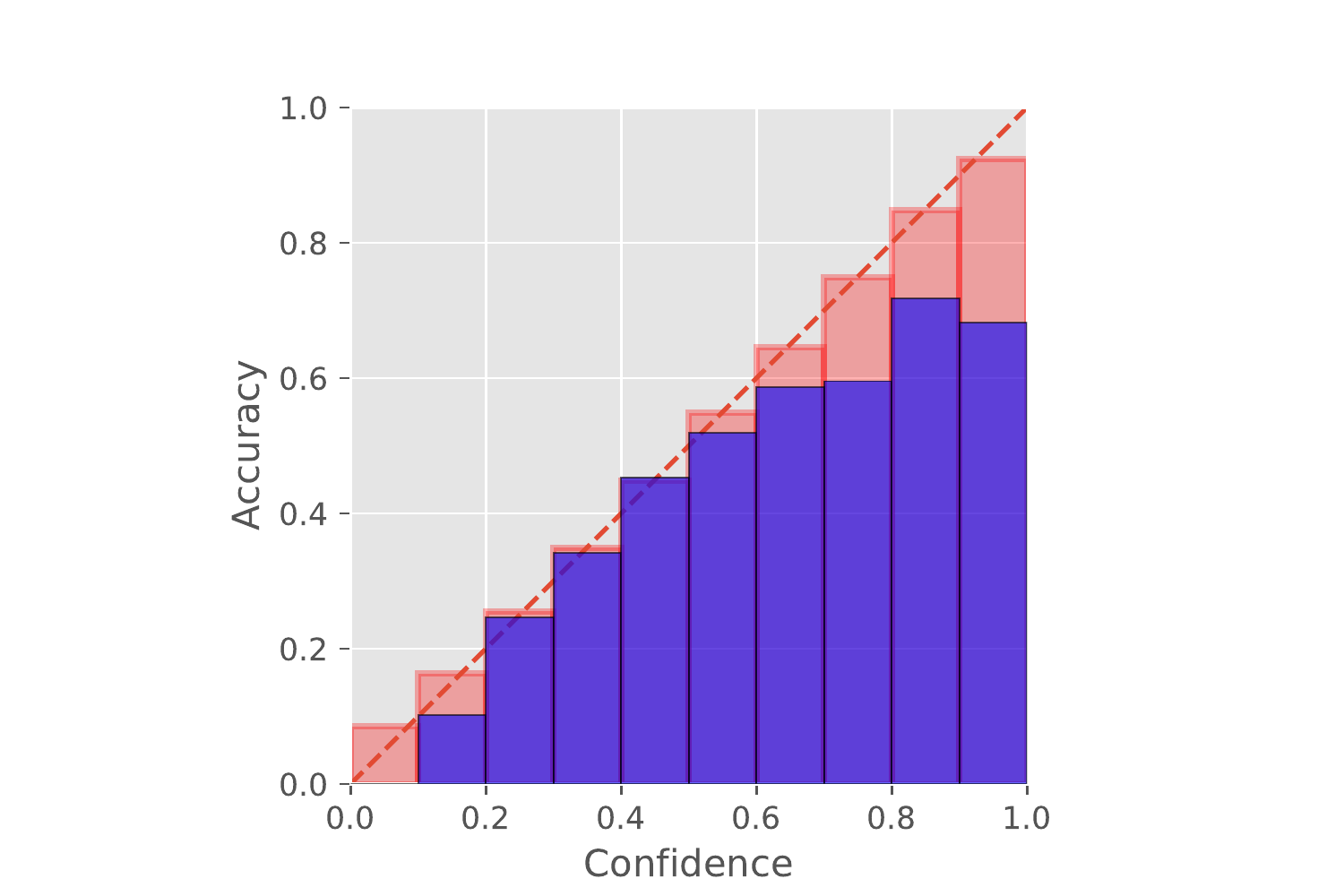}
  \end{subfigure}

  \caption{Reliability plots for uncalibrated  \textit{versus} \textsc{Individually} calibrated \textit{versus} \textsc{Jointly} calibrated on the \textsc{Generative QA} task using Temperature Scaling. \textbf{Blue bars} denote bin accuracy, \textbf{red bars} denote bin confidence, difference indicates miscalibration.} \label{fig:reliable}
\end{figure*}
 
We now present the results of the various calibration techniques in Table \ref{tab:main}. 
We also plot the reliability diagrams in Figure \ref{fig:reliable}. 
We compare all the described calibration algorithms in the three settings discussed.
As can be seen, in all the cases there is a benefit to \textsc{jointly} calibrate the \Ret and \Read. 
We give some reasons for why this setting works best in the discussion section below. 

\subsection{Discussion}
\subsubsection{Calibrating only the \Read}
In all our experiments we show that calibrating the \Read alone performs worse. We believe that this is because, at train time, the \Read is only trained on positive documents. This makes the \Read \textit{overconfident} on documents that don't have the answer. This phenomenon has been also been discussed in \citet{clark2017simple}.
We show an example in Fig \ref{fig:bad_reader}.
We also notice that adding the \Ret helps more in the QA task than for claim verification. 
We posit that this is because in the open-domain setting, the QA passage \Ret has a lower accuracy than the sentence \Ret for claim verification.\footnote{\texttt{QA, hits@10:0.77, CV, hits@5:0.94}\\ We use top-10 passages for QA and top-5 sentences for claim verification.}
\subsubsection{Calibrating \textsc{Individually}}

Our experimental results show that in almost all cases, it is detrimental to individually calibrate the \Read and \Ret. 
We believe that this is due to the \Ret's accuracy being misaligned with the final objective. 
In several cases, such as in QA, supervision for the \Ret is not provided, and instead a distant supervision objective is used where the document is marked as positive when it contains the answer string. 
We show an example in Figure \ref{fig:arch} where, for the question "Who won the women's worldcup in 2017", a document saying "world cup to be held in \textit{England}" would be assigned a positive label as it contains the answer string "England". This mismatch in accuracy for the \Ret can result in an incorrectly calibrated system. This problem has been well discussed in the literature and more recently by \citet{izacard2020distilling} 

\paragraph{Reliability plots}
As can be seen from Figure \ref{fig:reliable}, miscalibration results from the model being overconfident. This is evident with the blue bars being lower than the red -- model accuracy is less than model confidence for several bins. 
We also notice that all calibration techniques address this overconfidence by rescaling the output distribution.

\begin{table*}[t]
\centering
\small
%\vspace{-2mm}
\resizebox{\textwidth}{!}{%
\begin{tabular}{@{}l l  c c c c c c@{}}\toprule
\textbf{Task}  & \textbf{Setting}  & \textbf{Uncalibrated} & \textbf{Temp scaling}  & \textbf{Temp predictor} & \textbf{Forecaster}   \\ \midrule

\multirow{4}{*}{\textsc{Generative QA}} 
&  \textsc{Generator}  &  
  &  47.31
  & 45.22

 & 5.40

    \\
&  \textsc{Individually}  & 55.1 & 33.47    & 35.31  & 11.35    \\
&  \textsc{Jointly}  &  & {\bf 3.75}

 & {\bf 3.56} & {\bf 4.21}

  \\
  \midrule

\multirow{4}{*}{\textsc{Extractive QA}} 
&  \textsc{span extractor}  &  
  &  8.56  
 & 8.11
 & 4.68

    \\
&  \textsc{Individually}  & 37.1 & 10.32     & 7.42 & 12.74   \\
&  \textsc{Jointly}  &  &  \textbf{2.94}

 & \textbf{2.38} & \textbf{2.96}

  \\

\midrule

\multirow{4}{*}{\textsc{Claim Verification}}
& \textsc{Claim verifier}   & & 1.42
  &  1.64 & 1.66
 \\
&  \textsc{Individually}  & 7.02  & 16.35    & 23.6 &  26.73 \\
&  \textsc{jointly} &  &\textbf{ 1.15}
  & \textbf{1.30}& \textbf{0.98}  \\

\bottomrule
\end{tabular} 
}
% \vspace{-2mm}
\caption{Values in \% ECE, ($\downarrow$ is better). \textsc{\textbf{individually}} denotes the retriever and reader have been calibrated separately, while \textsc{\textbf{jointly}} indiciates that calibration on a joint objective.}
\label{tab:main}
\vspace{-2mm}
\end{table*}

\section{Analysis}
    \label{sec:analysis}
    Next, we attempt to verify the following claims:

\noindent {\bf C1:} The existing approach for calibrating only the reader doesn’t result in a good
calibration of the overall system. Jointly calibrating the reader and the retriever model is better.

\noindent {\bf C2:} Calibrated ODQA systems do better selective prediction when they are allowed to not provide answers to some questions.

\noindent {\bf C3:} Calibrated ODQA systems are better at handing domain shifts in questions at test time.

\noindent {\bf C4:} Calibrated ODQA systems are better at handling unanswerable questions at test time.

\subsection{Selective Prediction for Machine Reading}
One key use of confidence estimation is selective prediction. The selective prediction setting allows the model to decide whether it wants to make a prediction or abstain on each given test point. 
Selective prediction has been a long-standing research area in machine learning \cite{chow1957optimum, el2010foundations}. 

We investigate how different calibration methods perform on the task of selective prediction. 
There have been some recent efforts to understand selective prediction for QA models with regard to domain shift; \citet{kamath2020selective} investigate how forecasters can be effectively used as calibrators to predict when a model should abstain from providing an answer. 
We further this investigation in the open-domain setting to see if different calibration techniques can improve the model's performance on the selective prediction task. 
The evaluation metric used to judge a model's effectiveness in learning to abstain is the area under the risk-coverage curve.

Given an input $q$, the model's prediction $\hat a$ along with the confidence of the prediction $P(\hat a = a | q)$ and a threshold $\tau$, our model predicts the the answer $\hat a$ if $P(\hat a = a | q) \geq \tau$. For the test set and a value of $\tau$ there is an associated \textit{risk}: the fraction of the test set that the model answers incorrectly, and \textit{coverage}: the fraction of the test set the model makes a prediction on. As $\tau$ increases, so do the risk and coverage. We plot risk vs coverage as $\tau$ varies and report the area under the risk-coverage curve (\textsc{AURC}). Our results are shown in table \ref{tab:aurc}. 
We can infer from the results that all calibration methods help reduce the \textsc{AURC} to some extent however the Temperature predictor is able to perform the best on extractive QA while the forecaster is the best on claim verification and generative QA indicating that improving model calibration can also help for the task of selective prediction in the setting of machine reading.

\begin{table}
\centering
\small
%\vspace{-2mm}
\begin{tabular}{@{}l l  c c c @{}}\toprule
\textbf{Task}  & \textbf{Setting}  & \textbf{AURC}  \\ \midrule
\multirow{6}{*}{\textsc{Extractive QA}} &
 \textsc{Uncalibrated}  & 47.39  \\ &
\textsc{Forecaster}  & 44.21 
  \\ &
  \textsc{Temp scaling}  & 43.68 
  \\&
    \textsc{Temp prediction}  & 42.64 
  \\
  &\textsc{Best possible}  & 26.71
  \\
\midrule
\multirow{6}{*}{\textsc{Generative QA}} &
 \textsc{Uncalibrated}  &  53.57 \\ &
\textsc{Forecaster}  & 39.10
  \\ &
  \textsc{Temp scaling}  &  44.85
  \\&
    \textsc{Temp prediction}  &  43.21
  \\
  &\textsc{Best possible}  & 22.25
  \\
\midrule

\multirow{2}{*}{\textsc{Claim Verification}}
& \textsc{Uncalibrated}   & 11.04    \\
&  \textsc{Forecaster}  &  3.53   \\
&  \textsc{Temp scaling}  &  10.96   \\
& \textsc{Temp prediction} & 9.99 \\
& \textsc{Best possible} & 2.76 \\

\bottomrule
\end{tabular}
% \vspace{-2mm}
\caption{Area under Risk-Coverage curve. $\downarrow$ is better}
\label{tab:aurc}
\end{table}

\begin{figure}

    \raggedleft
    \includegraphics[width=0.99\linewidth, keepaspectratio]{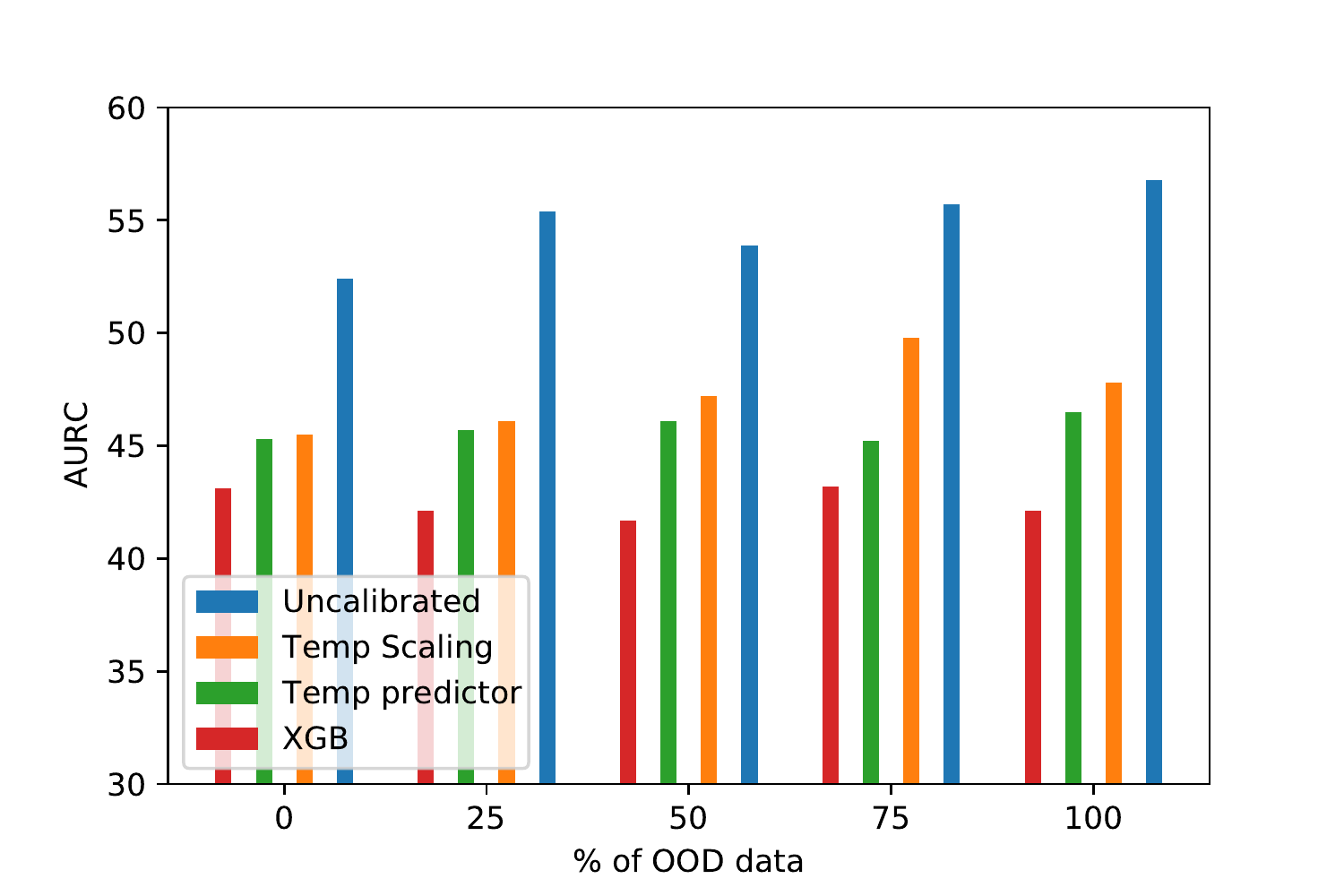}
    \caption{Area under risk coverage curves (AURC) using different calibration techniques}
    \label{fig:ood}
\end{figure}

\paragraph{Domain Adaptation}
With the increasing use of machine reading systems in the wild, a common problem encountered by them is that they are not resilient to inputs that do not come from the distribution of the data they were trained on. 
A method of selective prediction is often employed, where the model the model can abstain from answering the question.
\citep{kamath2020selective} show that training a seperate model to distinguish between in- and out-of- domain helps in doing selective prediction.
We show that a well calibrated model is able to perform better on the selective prediction setting even when the calibration step has no access to an out of distribution dataset. 
In our experiments, we calibrate a trained on NQ FiD model on the FiD dev set. 
We then evaluate the performance on different splits which contain varying percentages of out-of-distribution data (TriviaQA) \cite{joshi2017triviaqa}.
We plot the AURC with different splits containing different percentages of OOD questions in figure \ref{fig:ood}.
We notice that an uncalibrated model gets significantly worse when the amount of OOD samples are added. However calibation techniques are able to mitigate this and are able to maintain a steady AURC with increasing OOD samples. We found that the Forecaster (XGB) performing the best in this evaluation.

\begin{table*}[!htbp]
\small
\begin{tabular}{|p{60mm}|p{40mm}|p{50mm}|}
\hline
 Question & Retrieved passage and answer & Model confidences     \\ \hline
   \vspace{0.7pt} what harry potter movie came out in 2008 &  \vspace{0.3pt}  \dots \textbf{harry potter and the half - blood prince} is a 2009 fantasy film \dots     &   \textbf{Uncalibrated}: 
   $0.99 \times 0.79 = 0.78 $  \newline
   \textbf{Temp Scaling}:$0.98 \times 0.54 = 0.53$ \newline
   \textbf{Temp Prediction}:$0.85 \times 0.43 = 0.36 $ \newline
   \textbf{Forecaster}: $0.31$
   \\ \hline
    \vspace{0.7pt}who played the joker in the dark (k)night rises     & \vspace{0.3pt}  \dots he was played by australian actor\textbf{ heath ledger} \dots   &  
    \textbf{Uncalibrated}:$0.99 \times 0.57 = 0.56 $ \newline
   \textbf{Temp Scaling}: $0.98 \times 0.40 = 0.39$\newline
   \textbf{Temp Prediction}: $0.73 \times 0.33 = 0.24$\newline
   \textbf{Forecaster}:      $0.28$                                
       \\ \hline
   \vspace{0.7pt}who do you think you are book pdf &  \vspace{0.3pt}  Book of Ryan: \dots comedian cedric the \textbf{entertainer} makes a cameo \dots &   \textbf{Uncalibrated}: $0.97 \times 0.08 = 0.08$  \newline
   \textbf{Temp Scaling}:$0.78 \times 0.01 = 0.08$ \newline
   \textbf{Temp Prediction}:$0.63 \times 0.01 = 0.06$ \newline
   \textbf{Forecaster}: $0.03$ 
    \\ \hline
   \vspace{0.7pt}when it is winters in delhi how will the weather be in chennai &  \vspace{0.3pt} Chennai: Climate: \dots \textbf{coolest} part of the year is january \dots &     
   \textbf{Uncalibrated}:$0.99 \times 0.43 = 0.43$ \newline
   \textbf{Temp Scaling}: $0.96 \times 0.21 = 0.20$\newline
   \textbf{Temp Prediction}: $0.85 \times 0.21 = 0.18$\newline
   \textbf{Forecaster}:      $0.10$ 
    \\ \hline
   \vspace{0.7pt}how are the suburbs of paris different than those of most canadian cities & \vspace{0.3pt}  \dots Suburb: land use patterns in canadian suburbs are often \textbf{more mixed} \dots & 
   
    \textbf{Uncalibrated}:$0.98 \times 0.38 = 0.37$ \newline
   \textbf{Temp Scaling}: $0.96 \times 0.17 = 0.16$\newline
   \textbf{Temp Prediction}: $0.81 \times 0.19 = 0.15 $\newline
   \textbf{Forecaster}:      $0.07$ 
    \\ \hline

\end{tabular}
\caption{Examples of unanswerable questions. We show how each calibration approach is able to lower the confidence of the incorrect answer.}
\label{tab:unans}

\end{table*}

\subsection{Effect of $K$ on Calibration}
We analyze how the ECE values are affected by increasing the number of documents the reader model consumes. 
We run this experiment on the open-domain extractive QA task. 
 Along with $K=10$, we evaluate on $K=20$ and $K=50$.
As can be seen in Table \ref{tab:K_effect} all our methods scale well when we increase the value of $K$. 
Platt's scaling and temperature scaling are able to maintain low ECE scores even when the \Read collates the answer over multiple documents.
We believe that owing to the simplicity of the approach, these calibration methods are able to adapt well to different settings while in contrast, the forecaster and temperature predictor being more complex models struggle with this. 
The forecaster stands at a disadvantage here owing to the fact that the number of features given to the forecaster scales with $K$ and we believe that this could also have an impact on its inability to scale to larger values of $K$.

 \begin{table}[h]
 \centering
 \small
 \begin{tabular}{@{}l l  c c c @{}}\toprule
 \textbf{Calibration model}  & \textbf{$K=10$}  & \textbf{$K=20$}  & \textbf{$K=50$}  \\ \midrule

  \textsc{Temperature Scaling}  & 2.94 & 2.83 & 2.97\\ 
  \textsc{Platt's Scaling}  & 2.88 & 2.21 & 2.32\\ 
  \textsc{Temperature Predictor}  & 2.38 & 2.83 &  4.90\\ 
  \textsc{Forecaster}  & 2.96 & 4.62 & 5.10\\

 \bottomrule
 \end{tabular}
 \caption{How ECE changes with no of documents $K$}
 \label{tab:K_effect}
 \end{table}

\begin{figure}
    \raggedleft

    \includegraphics[width=0.95\linewidth, keepaspectratio]{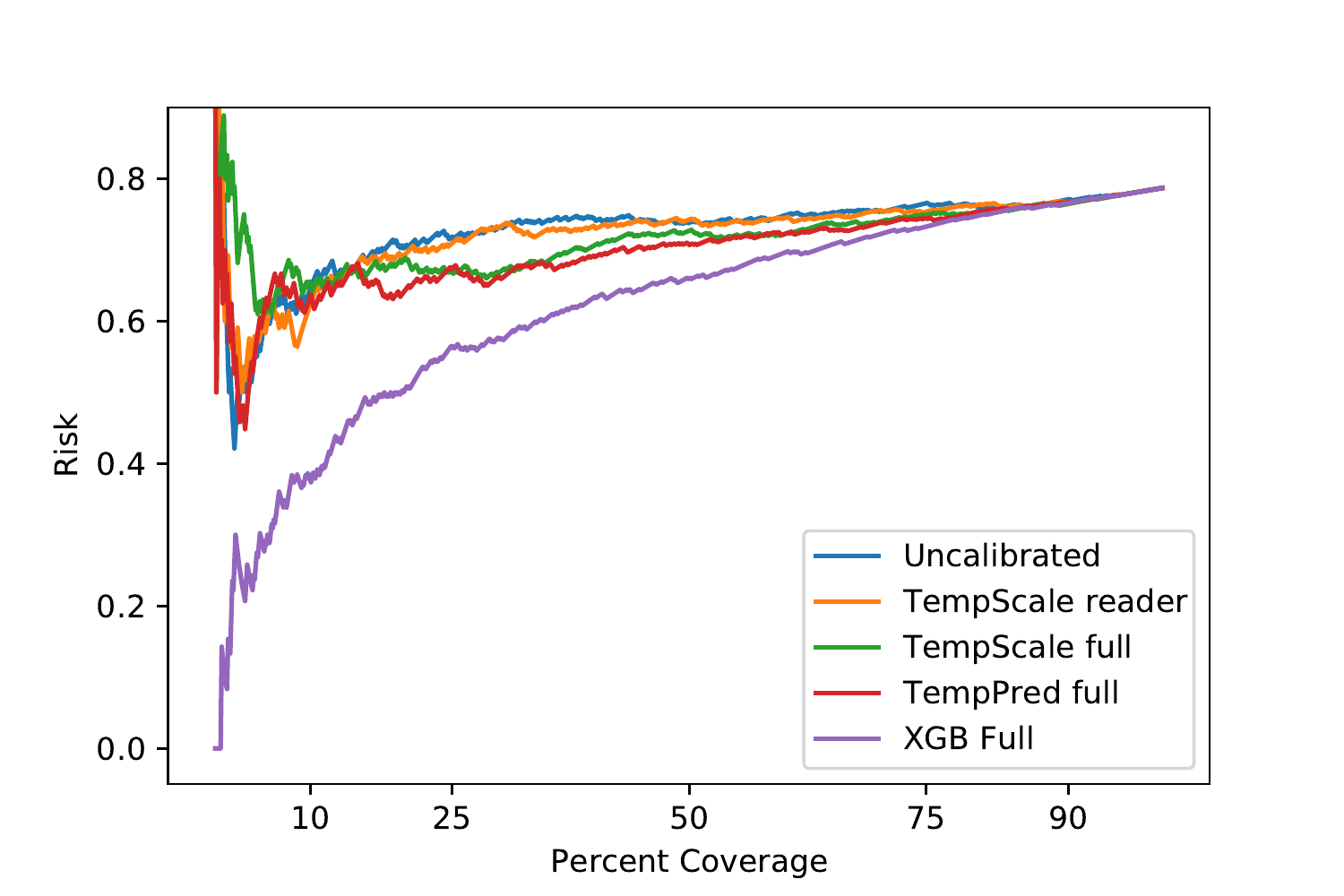}
    \caption{Risk coverage curve for unanswerable questions}
        \label{fig:unans}
\end{figure}

\begin{table*}[!htbp]
\small
\begin{tabular}{|p{80mm}|p{75mm}|}
\hline
 Question and Passage & Model confidences     \\ \hline
   \textbf{how many episodes  of corrie has there been}\newline 
   Clarkson (TV series):\dots The series ran for \textbf{ten} episodes, during a weekly airing schedule \dots
     &    
       \textbf{Uncalibrated}: $0.99 \times 0.49 = 0.49$  \newline
   \textbf{Temp Scaling}:$0.99 \times 0.23 = 0.22$ \newline
   \textbf{Temp Prediction}:$0.90 \times 0.18 = 0.16$ \newline
   \textbf{Forecaster}: $0.06$

  \\  
 \hline

       \textbf{what is in a pat o brien hurricane}\newline  
       Sucker hole: \dots Sucker hole is a colloquial term referring to a short spate of \textbf{good weather} \dots&    
       \textbf{Uncalibrated}: $0.99 \times 0.49 = 0.49$  \newline
   \textbf{Temp Scaling}:$0.99 \times 0.24 = 0.23$ \newline
   \textbf{Temp Prediction}:$0.95 \times 0.21 = 0.20$ \newline
   \textbf{Forecaster}: $0.07$
   \\
   \hline
\end{tabular}
\caption{Examples of questions where the \Ret fetches the wrong passages}
\label{tab:wrong_ret}
\end{table*}
\paragraph{Unanswerable Questions}
Another challenge that a user facing QA system can encouter is malformed questions. These include questions that were not probably questions, for example a user query containing a named entity which is a question or a question that cannot be answered because it contains a false premise. 
To investigate if a calibrated model can be used to abstain from answering such questions, we evaluate our approaches on the set of unanswerable questions proposed by \cite{asai2020challenges}.
We plot Risk-Coverage curves for different calibration techniques in Figure \ref{fig:unans}. 
We find all calibration techniques help in performing selective prediction when compared to an uncalibrated model. However, the Forecaster outperforms all other methods.
To exemplify how calibration techniques can help the model abstain we provide a few examples in table \ref{tab:unans}. 
These results come from running an extractive QA model on the set of unanswerable questions. 
We only evaluate using the top passage for simplicity. 
It can be seen that all calibration techniques are able to lower the confidence of the predicted answer in cases when the question is unasnwerable.
The first two questions provide examples of questions containing a false premise: No Harry Potter movie came out in 2008 and there was no Joker character in The Dark Knight Rises. 
In these cases, both the \Read and \Ret are confident about their prediction. 
The third question exemplifies a query that is not a question.
Here it can be seen that the \Read still places a high confidence in its provided answer, however, the \Ret assigns a very low score to its retrieved passage.
This provides a great insight into examples where using the retriever score can further help in calibration.
The last two questions are examples of types of questions generally not part of the dataset. They either require more sophisticated reasoning or are non-factoidal. 
Here we can observe that eventhough the \Read is confident, the retrieved passages are assigned a much lower score.
We find that by using the forecaster on generative QA we reduce the confidence of an incorrectly answered question by 46\% while decreasing the confidence of a correctly answered question by 38\%. 
\vspace{-0.1in}

\paragraph{\Ret mistakes}

Another common seen scenario in an open domain setting is when the \Ret is not able to provide any relevant passages.
In such cases, because the \Read is generally trained on only correct passages, it still produces a high confidence for the incorrect answer. We show that calibration methods that take into account the \Ret confidences can mitigate this by lowering the confidence of the answer. We provide few such examples in table \ref{tab:wrong_ret}

\section{Related Work}
    \label{sec:related}

Obtaining calibrated confidence scores for NLP tasks has recently gained attention. 
\citet{jagannatha2020calibrating} and \citet{jiang2020can} study how forecasters can be used and what features can be useful to calibrate the confidence of QA models. 
\citet{kamath2020selective} study calibration in the context of selective answering, i.e., learning when QA models should abstain from answering questions. 
They show that training a forecaster to predict the model's confidence can perform well when facing a distributional shift. 
\citet{su2019controlling} also investigate selective answering using a probe in the model to determine the model's confidence.

Also related to our work is \textit{uncertainity estimation} \cite{gal16,lakshminarayanan2017simple} as model uncertainities can be seen as confidence scores. In NLP, \newcite{Xiao_Wang_2019} propose an approach to characterize model and data uncertainties for various NLP problems. \newcite{Wang19} use uncertainty estimation for confidence estimation in MT. \newcite{dong-etal-2018-confidence} study confidence estimation for semantic parsing.
%In contrast, w
We are the first to study calibration of open-domain machine reading systems.

Our Gumbel-topk inspired approach of jointly calibrating the \Read and \Ret together is interesting in the light of recent open domain QA methods such as \citet{lewis2020retrieval} and \citet{sachan2021end} that train the entire system jointly.
As a future work we would also want to compare our approach with these other end-to-end training approaches for the task of calibration.

\section{Discussion and Conclusion}
    \label{sec:conclusion}
    In this paper, we analyzed how various calibration techniques can be adopted to open-domain machine reading systems which are now being used in user-facing scenarios. 
We showed that in such systems that include a retriever, calibrating the system's confidence is not trivial and we proposed a technique that allows calibration of the system \textit{jointly}.
Finally, we also provide an analysis on how the calibration techniques can help the model abstain from answering a question especially in settings where the model's prediction can be incorrect due to malformed or out-of-domain questions.

While we do not find evidence to prove that one calibration method (e.g. a gradient-based method) is better that the other (e.g. a forecaster approach), it would be important to investigate these questions with more nuanced human studies.
\section*{Ethical Considerations}
In recent years, deep learning approaches have been the main models of choice for practical machine reading systems. However, these systems are often overconfident in their predictions. A calibrated confidence score would help system users better understand the system's decision making. Our work introduces a simple and general way for calibrating these systems. While our models are not tuned for any specific application domain, our methods could be used in sensitive contexts such as legal or healthcare settings, and it is also essential that any work using our method undertake additional quality assurance and robustness testing before using it in their setting. The datasets used in our work do not contain any sensitive information to the best of our
knowledge.

\section*{Acknowledgments}
We gratefully acknowledge feedback and helpful comments from the various anonymous reviewers.
This work is partially supported by a Responsible AI grant by the Haslerstiftung and by an ETH Grant
(ETH-19 21-1).
SD is also partially supported by the IBM PhD fellowship.

\bibliographystyle{acl_natbib}
\bibliography{references}
\clearpage
\appendix

\section{Details on the Gumbel Top-K Solution}
\label{appendix:gumbel-topk}
The Gumbel-top-$K$ trick is a generalization the Gumbel–softmax. 
It essentially repeats the Gumbel trick $K$ times until we have a set of the desired size.

In order to sample a subset of size $K$ according to the categorical distribution given by $P(D_k| q)$, the method use the well-known two-step process to massage categorical sampling into a differentiable sampling procedure which includes: 1) reparameterization of the categorical using Gumbels and 2) softening the argmax into a softmax.  We formally describe the procedure below:

\paragraph{Gumbel top-K:}
 We first perturb the logits $S_{d_i}$ with Gumbel noise
 $n_i \sim \textit{Gumbel}(0; 1)$ such that $\tilde S_{d_i} = S_{d_i} + n_i$. Then, sampling from a categorical is equivalent to taking an argmax:
 \begin{align}
 d^* = \argmax_i \tilde S_{d_i}
 \end{align}
 
 In the top-K case, we start by sampling the first document using
 the gumbel pertubation and taking the argmax:
 \begin{align}
 d_1^* = \argmax_i \tilde S_{d_i}
 \end{align}
 Then we remove $d_1^*$ from the pool of documents under consideration and repeat the same procedure:
 \begin{align}
     d_2^* &= \argmax_{i \in \mathcal{D} \setminus \{d_1^*\}} \tilde S_{d_i} \\
  & \quad\quad\quad  \vdots  \nonumber \\
     d_k^* &= \!\!\!\!\!\!\!\!\!       \argmax_{i \in \mathcal{D} \setminus \{d_1^*, \ldots, d_{k-1}^*\}} \tilde S_{d_i} 
     \label{eq:hard_k_permute}
 \end{align}
 
 Now, we can construct a fully differentiable procedure by replacing the argmax with a softmax (the Gumbel softmax trick \cite{jang2017categorical}).
 We begin by relaxing the one-hot vector
 of the first document:
 \begin{equation}\label{eq:soft_k}
     d_i^{(1)} = \frac{\exp(\tilde S_{d_i}^{(1)})}{\sum_j \exp(\tilde S_{d_j}^{(1)})}
 \end{equation}
 Next, we continue relaxing the successive argmaxes with successive softmaxes \cite{plotz2018neural} as follows:
 \begin{align}
     d_i^{(2)} &= \frac{\exp(\tilde S_{d_i}^{(2)})}{\sum_j \exp(\tilde S_{d_j}^{(2)})} \\
  & \quad\quad\quad\quad\quad \vdots  \nonumber \\ 
    d_i^{(k)} &= \frac{\exp(\tilde S_{d_i}^{(k)})}{\sum_j \exp(\tilde S_{d_j}^{(k)})}
     \label{eq:soft_k_permute}
 \end{align}
 where we define the $\tilde S_{d_i}^{(k)}$ recursively 
 \begin{align}\label{eq:recursion}
     \tilde S_{d_i}^{(1)} &= \tilde S_{d_i} \\
     \tilde S_{d_i}^{(k)} &= \tilde S_{d_i}^{(k-1)} + \log\left(1 - d_i^{(k-1)}\right) &
 \end{align}
\newcite{xie2019reparameterizable} have shown that this procedure is a reasonable relaxation of the Gumbel-top-$K$. We refer the interested reader to their paper for more details.
 Finally, we sum over all the relaxed one-hot vectors $d_i^{(k)}$ 
to arrive at our softened $K$-hot retrieval:
 \begin{equation}\label{eq:final-gate}
     D_K = \sum_{k=1}^K d_i^{(k)}
 \end{equation}

This allows us to train the calibration parameters in $P(D_k | q)$ using the objective $\mathcal{L}_\theta$ .

\end{document}